\documentclass[sigconf]{acmart}
\AtBeginDocument{%
  \providecommand\BibTeX{{%
    \normalfont B\kern-0.5em{\scshape i\kern-0.25em b}\kern-0.8em\TeX}}}

\copyrightyear{2023} 
\acmYear{2023} 
\setcopyright{acmcopyright}\acmConference[WSDM '23]{Proceedings of the
Sixteenth ACM International Conference on Web Search and Data
Mining}{February 27-March 3, 2023}{Singapore, Singapore}
\acmBooktitle{Proceedings of the Sixteenth ACM International Conference on
Web Search and Data Mining (WSDM '23), February 27-March 3, 2023,
Singapore, Singapore}
\acmPrice{15.00}
\acmDOI{10.1145/3539597.3570418}
\acmISBN{978-1-4503-9407-9/23/02}

\usepackage{import}
\usepackage{CJKutf8}
\usepackage{rotating}
\usepackage{multirow}
\usepackage{makecell}

\newcommand\DS{Hansel}
\newcommand\TYDE{TyDE}

\newcommand{\entity}[1]{\texttt{#1}}
\newcommand{\ientity}[2]{\entity{#1}\textsuperscript{\href{https://www.wikidata.org/wiki/#2}{#2}}}
\newcommand{\eentity}[2]{{#1}\textsuperscript{\href{https://www.wikidata.org/wiki/#2}{#2}}}

\begin{document}

\title{Hansel: A Chinese Few-Shot and Zero-Shot 
Entity Linking Benchmark}


\author{Zhenran Xu}
\authornote{Both authors contributed equally to this research.}
\email{xuzhenran@stu.hit.edu.cn}
\orcid{0000-0002-5536-806X}
\affiliation{
  \institution{Harbin Institute of Technology}
  \city{Shenzhen}
  \country{China}
}

\author{Zifei Shan}
\email{zifeishan@tencent.com}
\orcid{0000-0002-8283-6498}
\authornotemark[1]
\affiliation{%
  \institution{Wechat, Tencent}
  \city{Shanghai}
  \country{China}}

\author{Yuxin Li}
\email{haidenli@tencent.com}
\orcid{0000-0001-7112-6952}
\affiliation{%
  \institution{Wechat, Tencent}
  \city{Shenzhen}
  \country{China}
}

\author{Baotian Hu}
\authornote{Corresponding author.}
\email{hubaotian@hit.edu.cn}
\orcid{0000-0001-7490-684X}
\affiliation{%
  \institution{Harbin Institute of Technology}
  \city{Shenzhen}
  \country{China}
}

\author{Bing Qin}
\email{qinb@ir.hit.edu.cn}
\orcid{0000-0002-2543-5604}
\affiliation{%
  \institution{Harbin Institute of Technology}
  \city{Harbin}
  \country{China}
}


\begin{abstract}
Modern Entity Linking (EL) systems entrench a popularity bias,
yet there is no dataset focusing on tail and emerging entities in languages other than English. 
We present Hansel, a new benchmark in Chinese that fills the vacancy of non-English few-shot and zero-shot EL challenges. 
The test set of Hansel is human annotated and reviewed, created with a novel method for collecting zero-shot EL datasets.
It covers 10K diverse documents in news, social media posts and other web articles, with Wikidata as its target Knowledge Base. We demonstrate that the existing state-of-the-art EL system performs poorly on Hansel (R@1 of 36.6\% on Few-Shot). We then establish a strong baseline that scores a R@1 of 46.2\% on Few-Shot and 76.6\% on Zero-Shot on our dataset.
We also show that our baseline achieves competitive results on TAC-KBP2015 Chinese Entity Linking task.
Datasets and codes are released at \url{https://github.com/HITsz-TMG/Hansel}.

\end{abstract}




\maketitle

\section{Introduction}
\label{sec:intro}
Entity Linking (EL) is the task of grounding a textual mention in context to a corresponding entity in a Knowledge Base (KB).
It is a fundamental component in applications such as Question Answering \cite{fevry2020entities, guu2020realm, de-cao-etal-2019-question}, KB Completion \cite{shen2014entity, zhang2014feature} and Dialogue \cite{curry2018alana}.

An unresolved challenge in EL is to accurately link against emerging and less popular entities. The \textit{Zero-Shot Entity Linking} problem was presented by \citet{zeshel}, aiming at linking mentions to entities unseen during training. 
On the other hand, \citet{ling2021} raised a common popularity bias in EL, i.e. EL systems significantly under-perform on tail entities that share names with popular entities.
Intuitively, we name the challenge to resolve tail entities as \textit{Few-Shot Entity Linking}, as most of them have only a few number of training examples.
Despite the aforementioned studies, non-English resources for zero-shot and few-shot EL are seldom available, hindering progress for these challenges across languages.

Moreover, existing zero-shot and few-shot EL datasets have a limited diversity,
because their collection methods rely on hyperlinks or manual templates.
\citet{zeshel} extracted mentions from Wikia articles hyperlinked to the Wikia KB, and
\citet{el100} used links from Wikinews to Wikipedia.
\citet{ling2021} generated AmbER sets by filling pre-defined templates with KB attributes.
These collection approaches are limited, as mentions are biased towards hyperlink editing conventions or syntactic templates.

To address the language bias and lack of syntactic diversity in few-shot and zero-shot EL datasets,
we present Hansel, a human-calibrated and challenging EL benchmark in simplified Chinese.
Hansel consists of few-shot and zero-shot test sets, with a Wikipedia-based training set.
The few-shot slice is collected from a multi-stage matching and annotation process. A core property of this slice is that all mentions are ambiguous and ``hard'' \cite{tsai-roth-2016-cross}, 
where the ground-truth entity is not the most popular by the mention.
The zero-shot slice is collected from a searching-based process: given a new entity's description, annotators find corresponding mentions and adversarial examples with Web search engines over diverse domains.
We demonstrate that both slices are challenging for state-of-the-art EL models. 
We further design a type system based on rich Wikidata structure, and propose a novel architecture utilizing the type system that improves over dual-encoder based models.

\begin{table*}[!ht]
\centering
\begin{tabular}{r rrr rrr rrr} \toprule
  & \multicolumn{3}{c}{\bf \# Mentions} 
  & \multicolumn{3}{c}{\bf \# Documents} 
  & \multicolumn{3}{c}{\bf \# Entities} 
  \\ 
\cmidrule(lr){2-4}  \cmidrule(lr){5-7}  \cmidrule(lr){8-10}  
 & In-KB & NIL & Total & In-KB & NIL & Total & $E_{known}$ & $E_{new}$   & Total \\ 
\midrule
Train & 9.88M & - & 9.88M & 1.05M & - & 1.05M  & 541K & -     & 541K \\
Validation & 9,674 & - & 9,674 & 1,000 & - & 1,000  & 6,320 & -     & 6,320 \\
Hansel-FS & 3,404 & 1,856 & 5,260 &  3,389 & 1,850 & 5,234  & 2,720 & -     & 2,720
\\
Hansel-ZS & 4,208 & 507   & 4,715 &  4,200 & 507  &  4,704  & 1,054 & 2,992 & 4,046 \\
\bottomrule
\end{tabular}
\caption{
\small \textbf{Statistics of the \DS \space dataset.}
We break down the number of mentions and documents by whether the label is a NIL entity or inside Wikidata (In-KB),
and the number of distinct entities by whether the entity is in an emerging entity in $E_{new}$.
\label{tab:stats_hansel}}
\end{table*}

The main contributions of this work are:
\begin{itemize}
\itemsep0em
\item Publish Hansel, a challenging multi-domain benchmark for Chinese EL with Wikidata as KB, featuring a zero-shot slice with emerging entities, a few-shot slice with hard mentions, and a large training set with 1M documents.
\item  Propose a novel and feasible zero-shot entity linking dataset collection method, applicable for any language.
\item Achieve strong results on TAC-KBP2015 Chinese EL task with a monolingual model, on a par with state-of-the-art multilingual models on this task.

\end{itemize}

\section{Related Work}

For years, the primary focus of entity linking studies were constrained to English-only and fixed-KB \cite{ling2015design,fevry2020empirical,relic,genre}.
Cross-lingual entity linking was introduced to link non-English mentions to English KBs  \cite{mcnamee-etal-2011-cross,ji2015overview}.
Recently, \citet{el100} introduced \textbf{Multilingual Entity Linking}, a more general formulation to link mentions from any language to a language-agnostic KB. 
Their Mewsli-9 multilingual benchmark alleviates the language bias in general EL to some extent, but many languages including Chinese are not covered.

\textbf{Zero-Shot Entity Linking} was proposed by \citet{zeshel}, 
with an English zero-shot EL dataset.
Mewsli-9 has a zero-shot slice of 3,198 multilingual mentions, but only contains Wikinews hyperlinks.
Zero-shot EL on temporally evolving KBs is less discussed.
To this end, \citet{hoffart2014discovering} proposed EL on emerging entities, but the dataset is English-only.
In this work, we present the first non-English zero-shot EL dataset on emerging entities.

\textbf{Few-Shot Entity Linking} was frequently studied in recent years. \citet{provatorova-2021-robustness} suggested that 
high accuracy on previous EL datasets can be obtained by merely learning the prior, and released ShadowLink test set whose ``Shadow'' subset is similar to our few-shot setting, but only available in English.
\citet{ling2021} discovered that current EL systems significantly under-perform on tail entities, and released AmbER test sets. Their dataset is English-only and generated by filling pre-defined templates with KB attributes.
\citet{tsai-roth-2016-cross} has a cross-lingual ``hard'' subset similar to our setting, but only contains Wikipedia hyperlinks.
In this work, we present the first non-English, human-calibrated few-shot EL dataset with better syntactic diversity.

In Chinese language, existing EL datasets are very limited.
An established dataset is TAC-KBP2015~\cite{ji2015overview}.
DuEL \cite{ccks2019} is an EL dataset annotated to an incomplete subset of Baidu's knowledge base (390K entities) and thus cannot serve as a comprehensive EL benchmark.
None of the above datasets focus on zero-shot or few-shot EL.
More comparison of these datasets and their limitations are discussed in Section~\ref{sec:compare_discuss}.
Our proposed benchmark enriches Chinese EL resources and alleviates their popularity bias, providing basis for Chinese and multilingual few-shot and zero-shot EL studies.

\begin{figure*}[ht]
\centering
\includegraphics[width=\textwidth]{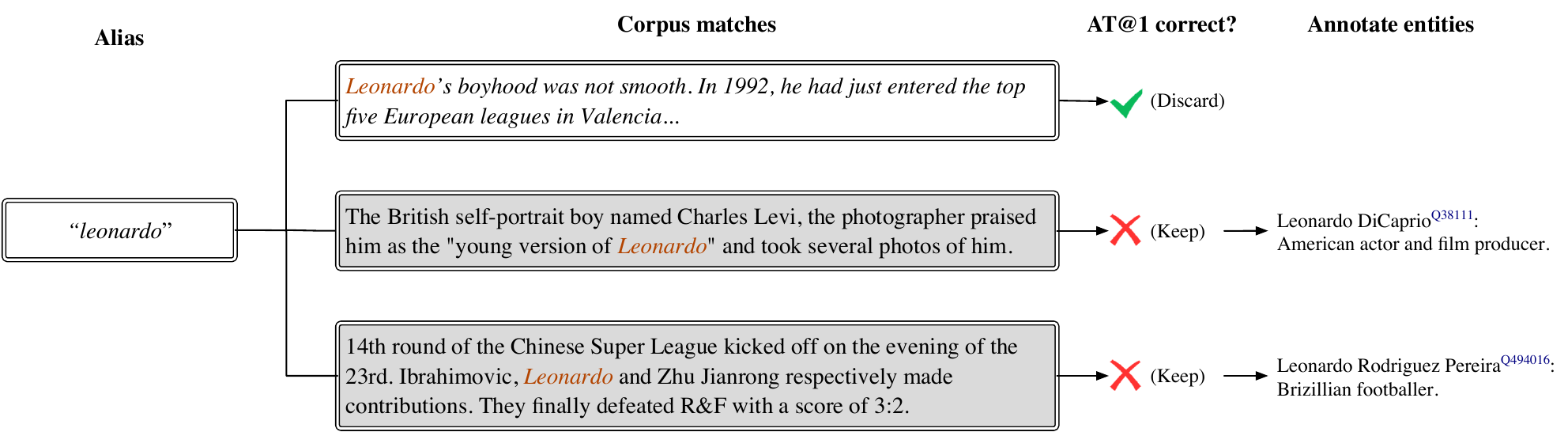}
 \caption{
  \small \textbf{Annotation process for the Few-Shot dataset}, with a translated example in Hansel-FS. We first match aliases against the corpora to generate diversified potential mentions, then annotate if the most popular entity (AT@1) is the correct candidate for each mention. We only keep cases where AT@1 is incorrect, and then annotate the correct entity.
 \label{fig:annotation_fs}
   } 
\end{figure*}

\begin{figure*}[t!]
\centering
\includegraphics[width=\textwidth]{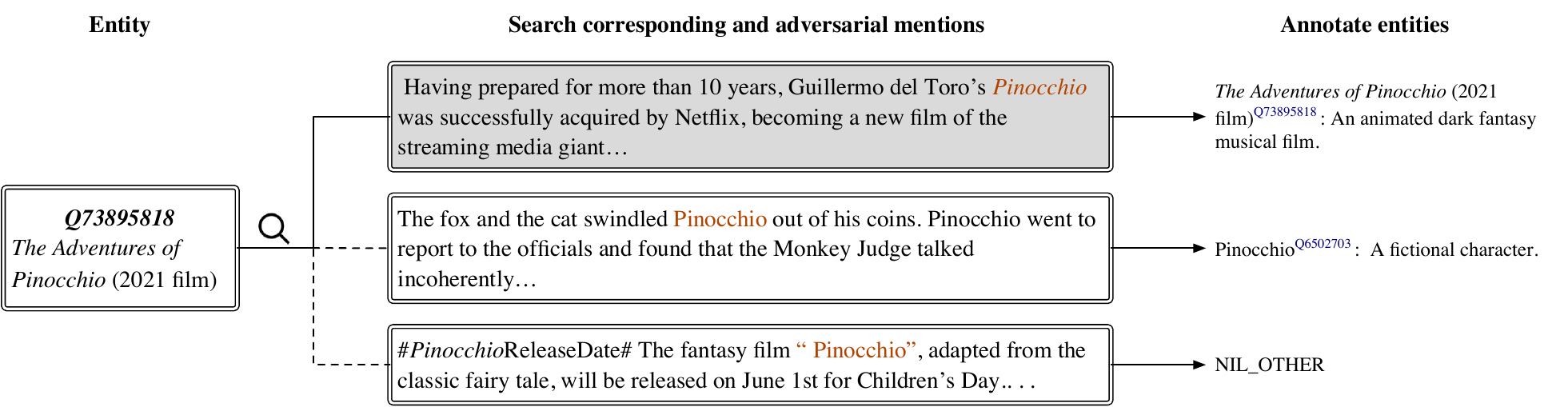}
 \caption{
   \small \textbf{Annotation process for the Zero-Shot dataset}, with a translated example in Hansel-ZS. Given a new entity, we search on the Web for a corresponding mention, and a few mentions that share the same mention text but refer to different entities.
\label{fig:annotation_zs}} 
\end{figure*}

\section{Hansel Dataset}

Define entries in a Knowledge Base (KB) as a set of \textbf{entities} $E$. Given an input document $D = \{s_1, \dots, s_d\}$ and \textbf{entity mentions} that are spans with known boundaries: $M_D = \{m_1, \dots, m_n\}$, an entity linking (EL) system outputs mention-entity pairs: $\{(m_i, e_i)\}_{i \in [1,n] }$, where each entity is either a known KB entity or NIL (i.e., an entity outside KB): $e \in E \cup \{nil\}$.  
Another setting of EL where mention spans are not given \cite{genre} is out of scope for this work.

We publish an EL dataset for simplified Chinese (zh-hans), named Hansel. The training set is processed from Wikipedia. The test set of Hansel contains Few-Shot (FS) and Zero-Shot (ZS) slices, focusing respectively on tail entities and emerging entities. Both test sets contain mentions from diverse documents, with the ground truth entity ID annotated. Dataset statistics are shown in \autoref{tab:stats_hansel}.

\subsection{Knowledge Base}

To reflect temporal evolution of the knowledge base, we split Wikidata entities into \textbf{Known} and \textbf{New} sets using two historical dumps: 

\textbf{Known Entities ($E_{known}$)} refer to Wikidata entities in 2018-08-13 dump.
All our models are trained with $E_{known}$ as KB.

\textbf{New Entities ($E_{new}$)} refer to Wikidata entities in 2021-03-15 dump that do not exist in $E_{known}$. Intuitively, entities in $E_{new}$ were newly added to Wikidata between 2018 and 2021 thus never seen when training on 2018 data, thus considered as a zero-shot setting.

\textbf{Entity filtering.} We filter Wikidata entities to get a clean KB:
we remove all instances of disambiguation pages, templates, categories, modules, list pages, project pages, Wikidata properties, as well as their subclasses.
For the scope of our work, we further constrain to entities with Chinese Wikipedia pages. 
After filtering, there are roughly 1M entities in $E_{known}$ and 57K entities in $E_{new}$.

\textbf{Alias table.} 
An alias table defines the prior probability of a text mention $m$ linking to an entity $e$, 
i.e. $P(e|m)$, estimated as follows:
\begin{equation}\label{eq:prior}
    P(e|m) = \frac{count(m,e)}{count(m)},
\end{equation}
where $count(m)$ denotes the number of anchor texts with the surface form $m$ in Wikipedia;
$count(m,e)$ denotes the number of anchor texts with the surface form $m$ pointing to the entity $e$.
We extract an alias table \textit{AT-base} from Wikipedia 
2021-03-01 by parsing Wikipedia internal links, redirections and page titles.

\label{sec:type_system}
Prior work showed that types can benefit EL systems \cite{ling2015design,deeptype,tabi-2022-acl}.
We present a new formulation for coarse and fine entity typing, utilizing rich structural knowledge in Wikidata:

\textbf{Coarse Types.}  Define Wikidata entities as $E$, Wikidata property types as $P$, and relation triples as $R(e_1, p, e_2)$. We define a transitive typing feature denoted as $Type$: 
\begin{gather*}\label{eq:transitive1}
R(e_1, P31, e_2) \Rightarrow Type(e_1, e_2) \\
Type(e_1, e_2) \wedge R(e_2, P279, e_3) \Rightarrow Type(e_1, e_3)
\end{gather*}
where $P31$ stands for \emph{instance of} and $P279$ for \emph{subclass of}.

\begin{table}[t]
\small
\centering
\begin{tabular}{r l} \toprule
\textbf{Coarse Type} & \textbf{Definition} \\ \midrule
$PER(e)$ & $Type(e, Q215627)$     \\
$LOC(e)$ & $Type(e, Q618123)$ \\
$ORG(e)$ & $Type(e, Q43229)$ \\
$EVENT(e)$ & $Type(e, Q1656682)$ \\
$OTHER(e)$ & All other entities \\
\bottomrule
\end{tabular}
\caption{Coarse types defined with transitive $Type$.
\label{tab:coarse_types}}
\end{table}

\begin{table}[t]
\small
\centering
\begin{tabular}{l l} \toprule
{\bf    TopSnak} & {\bf Definition} \\ \midrule
 P31-Q13442814   & instance of: scholarly article \\
 P17-Q148        & country: People's Republic of China \\
 P21-Q6581072    & sex or gender: female \\
 P106-Q82955     & occupation: politician \\
 P641-Q2736      & sport: association football \\
\bottomrule
\end{tabular}
\caption{Examples of TopSnaks.
\label{tab:5topsnaks}}
\end{table}

We define five categories with the above feature in~\autoref{tab:coarse_types}: person (PER), location (LOC), organization (ORG), event (EVENT), and others (OTHER). Note that our LOC combines GPE, LOC and FAC types as defined in ACE \cite{doddington2004automatic} to better fit Wikidata typing guideline\footnote{We refer to
    \url{https://www.wikidata.org/wiki/Wikidata:WikiProject_Infoboxes}
    when choosing appropriate entities for corresponding types.}. 

\textbf{Fine Types.} Our fine typing system \emph{TopSnaks} is defined as top 10,000 property-value pairs, i.e. $(p, e_{2})$ tuples, sorted by frequency in KB\footnote{``SNAK'' is a Wikidata term referring to ``some notation about knowledge'' : \url{https://www.wikidata.org/wiki/Q86719099}.}. 
As the 5 examples of Wikidata TopSnaks in \autoref{tab:5topsnaks} show, TopSnaks include diverse entity attributes such as types, gender, occupation, country and sport. 
We verify that the TopSnaks generated from the 2018 Wikidata dump covers about 90\% of $E_{new}$, indicating good generalization over emerging entities.

\subsection{Training Data}
\label{sec:dataset_train}

Following previous work \cite{el100,genre}, we use Wikipedia internal links to construct a training set. The alignment of Wikidata and Wikipedia ecosystems enables utility of rich hyperlink structures in Wikipedia.

All new entities $E_{new}$ are kept unseen during training. 
Ideally, one would acquire the 2018 Wikipedia dump as the training corpus. 
As the full 2018 Wikipedia dump is not publicly available, we use 2021-03-01 Wikipedia dump and hold out all entity pages mapped to $E_{new}$ as well as all mentions with pagelinks to $E_{new}$ entities. 
The training set contains 9.9M mentions from 1.1M documents.
We hold out 1K full documents (9.7K mentions) as the validation set.

\subsection{Few-Shot Evaluation Slice}
\label{sec:dataset_fewshot}

For the Few-Shot (FS) test set, we collect human annotations 
in three Chinese corpora: 
LCSTS \cite{hu-etal-2015-lcsts}, covering short microblogging texts,
SohuNews and TenSiteNews \cite{wang2008automatic}, covering long news articles.

\textbf{Matching.} The FS slice is collected based on a matching-based process as illustrated in \autoref{fig:annotation_fs}. We first use \textit{AT-base} to match against the corpora to generate potential mentions, then randomly sample for human annotation.
Note that we only match ambiguous mentions with at least two entity candidates in $E_{known}$, and keep limited examples per mention word for better diversity.

\textbf{Annotation.} Human annotation was performed on more than 15K examples with 15 annotators. For each example, annotators first modify the incorrect mention boundary, or remove the example if it is not an entity mention. 
Then, they select the referred entity from candidate entities given by \textit{AT-base}. 
For each candidate, annotators have access to its description (first paragraph in Wikipedia) and Wikipedia link. 
If the candidate with the highest prior (AT@1) is correct, then the example is discarded. 75\% of examples are dropped in this step.
If none of the candidates are correct, the annotator find the correct Wikipedia page
for the entity through search engines.
If no Wikipedia page can be found, they label a NIL entity with its coarse type from~\autoref{tab:coarse_types}.
\autoref{fig:example_known} shows an example of the FS slice.

\begin{table*}[!ht]
\small
\centering
\begin{tabular}{rp{.85\textwidth}}
\toprule
  
  \textbf{Context} & 
  \begin{CJK}{UTF8}{gbsn} 
  9月23日，《 [E1] 雷雨 [/E1] 》作为北京人艺纪念曹禺先生诞辰百年的压轴之作，即将再登首都剧场 \end{CJK} \ldots  \\ 
  \cmidrule(lr){2-2}

\textbf{Translation} &
On September 23, ``\textbf{[E1] Thunderstorm [/E1]}'', as the finale of Beijing Renyi's commemoration of the centenary of Mr. Cao Yu's birth, will soon appear at the Capital Theater.
  \ldots  \\  \cmidrule(lr){2-2}
  
  \textbf{Annotation} & \ientity{\begin{CJK}{UTF8}{gbsn}雷雨\_(话剧)\end{CJK}}{Q5372480}: \begin{CJK}{UTF8}{gbsn}	《雷雨》是中国现代剧作家曹禺的处女作 \end{CJK} \ldots  \\ \cmidrule(lr){2-2}
  
  \textbf{Translation} &
  \eentity{Thunderstorm\_(play)}{Q5372480}: Thunderstorm is the debut work of modern Chinese playwright Cao Yu \ldots  \\ 

\bottomrule
\end{tabular}
\caption{
An Example in \DS-FS.
\label{fig:example_known}}
\end{table*}

\subsection{Zero-shot Evaluation Slice}
\label{sec:dataset_zeroshot}

Collecting a Zero-Shot (ZS) slice is challenging, due to the difficulty
to find occurrences of new entities on a fixed text corpus, especially when the corpus has no hyperlink structure. To address this challenge, we design a novel data collection scheme by searching entity mentions across the Web given an entity description.

\textbf{Type balancing.} We first sample from $E_{new}$ to get a subset with diversified coarse types, as the original type distribution of $E_{new}$ is heavily biased towards OTHER (52\%) and PER (38\%). We sample from $E_{new}$ by 50\% random sampling and 50\% uniform sampling.

\textbf{Searching-based Annotation.} 
Given the title, description and aliases of an entity in $E_{new}$,
annotators search the Internet\footnote{To facilitate searching, we provide annotators with pre-filled search query templates in an annotation tool, such as Google queries with entity names and target domains.} 
for a corresponding mention and collect the mention context. They further seek 1 or 2 adversarial examples by searching for a same or similar mention referring to a different entity. 
The process is shown in \autoref{fig:annotation_zs} with an annotation example.
Such ambiguous mentions introduce more diversity on this dataset.
\autoref{fig:example_zs} shows another example and its adversarial mention in the ZS slice.

\begin{table*}[!ht]
\small
\centering
\begin{tabular}{rp{.85\textwidth}}
\toprule
  \textbf{Mention 1} & 
  \begin{CJK}{UTF8}{gbsn} 2019年 [E1] 上海大师赛 [/E1] 举行了男单正赛的抽签仪式。今年进入网球名人堂的李娜与获得男单正赛外卡的张之臻\end{CJK} \ldots  \\  \cmidrule(lr){2-2}

  \textbf{Translation} &
  The draw of men's singles competition was held in 2019 \textbf{[E1] Shanghai Masters [/E1]}. Na Li, who entered the Tennis Hall of Fame  \ldots  \\  \cmidrule(lr){2-2}
  
  \textbf{Entity 1} & \eentity{\begin{CJK}{UTF8}{gbsn}2019年上海大师赛\end{CJK}}{Q69355546}: \begin{CJK}{UTF8}{gbsn}2019年上海大师赛为第12届上海大师赛，是ATP世界巡回赛1000大师赛事的其中一站\end{CJK} \ldots  \\ \cmidrule(lr){2-2}
  
  \textbf{Translation} & \eentity{\begin{CJK}{UTF8}{gbsn}2019 Shanghai Masters\end{CJK}}{Q69355546}:
  The 2019 Shanghai Masters was the 12th edition of the Shanghai ATP Masters 1000
  \ldots  \\  \midrule

  \textbf{Mention 2} & 
  \begin{CJK}{UTF8}{gbsn} \#2020斯诺克世锦赛\# 交手记录 \ldots 
  2019年 [E1] 上海大师赛 [/E1] 半决赛：奥沙利文10-6威尔逊
  \end{CJK} \ldots  \\  \cmidrule(lr){2-2}
  
  \textbf{Translation} &
  \#2020 World Snooker Championship\# Match Record \ldots
  2019 \textbf{[E1] Shanghai Masters [/E1]} Semi-final: O'Sullivan 10-6 Wilson 
  \ldots  \\  \cmidrule(lr){2-2}
  
  \textbf{Entity 2} & \eentity{\begin{CJK}{UTF8}{gbsn}2019年斯诺克上海大师赛\end{CJK}}{Q66436641}: \begin{CJK}{UTF8}{gbsn}
  2019年世界斯诺克·上海大师赛属于2019年9月9日－15日在上海富豪环球东亚酒店举行
  \end{CJK} \ldots  \\ \cmidrule(lr){2-2}
  
  \textbf{Translation} & \eentity{\begin{CJK}{UTF8}{gbsn}2019 Shanghai Snooker Masters\end{CJK}}{Q66436641}:
  The 2019 World Snooker Shanghai Masters took place at the Regal International 
\ldots  \\  \midrule
  
\textbf{Analysis} & During data collection, Entity 1 (entity in $E_{new}$) was provided. The annotator found Mention 1 via Web search, as well as an adversarial Mention 2 with the same phrase ("Shanghai Masters"), referring to a tennis tournament and a snooker tournament respectively.
  \\

\bottomrule
\end{tabular}
\caption{
An example and its adversarial mention collected by annotators in \DS-ZS.
\label{fig:example_zs}}
\end{table*}

\subsection{Dataset Quality and Statistics}

\textbf{Expert checking.} For both FS and ZS slices, after the first pass of annotation, there is an expert-checking phase, where 5 human experts manually examine and correct all annotated examples.
``Experts'' are well-trained annotators who made fewest mistakes in the trial annotation and learned basic knowledge of entity linking.
Each example is labeled by one annotator and reviewed by one expert (i.e. tie-breaking by choosing the expert's result).
The expert-reviewed results are used as the ground truth (GT) of this dataset.

\textbf{Dataset statistics.} As \autoref{tab:stats_hansel} shows, the FS slice has 5,260 mentions from 5,234 documents, covering 2,720 diverse entities. The ZS slice has 4,715 mentions across 4,707 documents, covering 4,046 distinct entities.
Domains are news (51.9\%) and social media (48.1\%) for FS slice, and news (38.6\%), social media (14.9\%), and other articles such as E-books and commerce (46.5\%) for ZS slice.

\textbf{Dataset Quality.}
To measure dataset quality, we first calculate the percentage agreement between the annotator and the expert.
The percentage agreement of Hansel-FS and Hansel-ZS are 87.3\% and 95.9\% respectively, i.e. modification rate is 12.7\% and 4.1\% during expert checking.
Both imperfect mention boundaries and wrong entities count as disagreements. Boundary changes account for 40.1\% for FS disagreements and 53\% for ZS.

We further take a random sample from the final dataset, 100 examples from FS and 100 from ZS.
Two annotators independently label whether the GT entity is correct. 
In this step, two annotators agree on 88\% of the cases in FS slice and 94\% of the cases in ZS slice.
We use Cohen's Kappa coefficient to evaluate the inter-annotator agreement. The coefficient is 0.622 for FS and 0.651 for ZS, indicative of substantial agreement between annotators \cite{fleiss1973equivalence}.
Average human accuracy (evaluating on GT) is 88\% for FS and 95.5\% for ZS.

\section{Models}

\begin{figure*}[ht!]
\centering
\includegraphics[width=\textwidth]{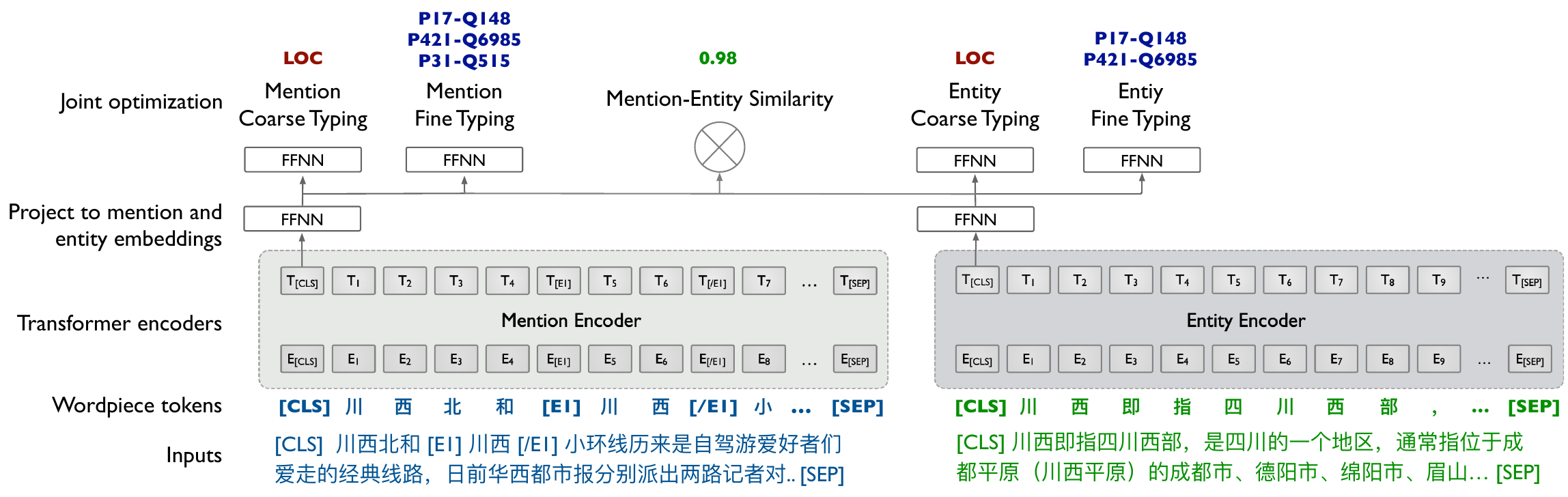}
 \caption{ 
  \small \textbf{Typing-enhanced Dual Encoder (\TYDE)} architecture. Both mention and entity encoders are 12-layer transformer encoders
initialized from BERT-base,
projecting mention in context (annotated with [E1] and [/E1] markers) and entity description to 256-d embeddings. 
Cosine similarity between mention and entity embeddings is jointly optimized with typing losses.
\label{fig:architecture}} 
\end{figure*}

We establish baselines on Hansel with a Dual Encoder (DE) model and a Cross-Attention encoder (CA) model for entity disambiguation. 
We also present a novel architecture that exploit our coarse and fine typing system to add typing-based supervision on DE.

\subsection{Dual Encoder Model}

Following previous work \cite{wu2019scalable,el100}, we train a Dual Encoder (DE) model
to project entity and mention contextual representations into a same vector space. 
Such models are scalable in that the entity embeddings can be pre-computed and stored, enabling fast 
retrieval or dot-product based similarity scoring.

The dual encoder takes a mention-entity pair $(m,e)$ and outputs their cosine similarity score:
\begin{equation}\label{eq:cosine}
    sim(m,e) = \frac{\phi(m)^T \psi(e)}{\|\phi(m)\| \|\psi(e)\|},
\end{equation}
where both $\phi$ and $\psi$ are learned transformer encoders projecting mention and entity input sequences into $d$-dimensional vectors ($d$=256),
i.e. mapping the $[CLS]$ token with a dense layer to the output embedding. 
We use boundary tokens (denoted as $[E1]$ and $[/E1]$) to wrap mentions for the input of $\phi$. 
We concatenate the title and the description as an entity's description for the input of $\psi$. 
The DE model is optimized with in-batch sampled softmax loss.

We use the DE model as a scoring step on candidates generated by the alias table \textit{AT-base}, combining the model's prediction $sim(m,e)$ with the prior $P(e|m)$ to produce a score $s(m,e)$:
\begin{equation}\label{eq:priorcosine}
    s(m,e) = P(e|m)sim(m,e).
\end{equation}

\subsection{Cross-Attention Encoder Model}

Following~\citet{wu2019scalable}, Cross-Attention encoder (CA) takes concatenated mention and entity as input and outputs their similarity (in the range 0 to 1), optimized with a binary cross-entropy loss.

Since the training set only comes with positive examples, 
we collect incorrect entities retrieved by the alias table as negative examples, 
and randomly keep 20\% of them to reduce label imbalance.

\subsection{TyDE: Typing-enhanced Dual Encoder}
\label{sec:modeling_tyde}

Previous work \cite{ling2015design,deeptype} suggested that type coherence can benefit EL systems.
However, models like DE or CA only implicitly learn type coherence with pretrained contextualized representations. Moreover, types for new entities in KB can be incomplete.

We propose a novel typing-enhanced dual encoder (TyDE), 
using type prediction as an auxiliary supervision task to improve the dual encoder.
As \autoref{fig:architecture} shows,
on top of mention and entity encodings output by $\phi$ and $\psi$, 
we add classification layers for coarse and fine typing. 
On each side, we use a softmax classifier for coarse types and a binary classifier for each of 10K fine types.
The TyDE model is optimized with type classification losses in addition to in-batch sampled softmax loss.
The supervision approach does not rely on types as encoder input, thus less affected by KB incompleteness. 

During inference, we use the similarity score as defined in DE, $P(e|m)sim(m,e)$,
and combine it with the predicted coarse and fine typing scores. 
Note that we do not require entity types for inference.
Coarse typing score $S_{c}$ and fine typing score $S_{f}$ are defined as:
\begin{equation}\label{eq:type1}\begin{split}
    s_{c}(m,e) = \sigma_{c}(m)^T \rho_{c}(e), \\
    s_{f}(m,e) = \sigma_{f}(m)^T \rho_{f}(e)
\end{split}\end{equation}
where $\sigma_{c}$, $\rho_{c}$, $\sigma_{f}$ and $\rho_{f}$ are single linear dense layers, projecting $\phi$ and $\psi$ outputs to corresponding type dimensions. $\sigma_{c}$ and $\rho_{c}$ project to 5 coarse types, and $\sigma_{f}$ and $\rho_{f}$ project to 10,000 fine types.

There are 2 different scoring settings for TyDE:
(1) similarity only, i.e. $P(e|m)sim(m,e)$, so typing information is only used implicitly via co-training;
(2) multiply similarity with coarse, fine, or both typing scores, i.e. $P(e|m)sim(m,e)s_{c}(m,e)$, $P(e|m)sim(m,e)s_{f}(m,e)$, $P(e|m)sim(m,e)s_{c}(m,e)s_{f}(m,e)$ respectively.
Note that the combination requires trivial additional computation for scoring.
Evaluation of these settings is detailed in Section \ref{sec:eval_tackbp2015}.

\section{Experiments}
In this section, we first describe implementation details (Section \ref{sec:experiments}).
Then we evaluate our baseline models on TAC-KBP2015, an established Chinese dataset with the most reported results, to show our model's competitive result with the state-of-the-art mGENRE \cite{mgenre} (Section \ref{sec:eval_tackbp2015}).
Next, we set baselines on Hansel with our models and mGENRE (Section \ref{sec:eval_hansel}), 
showing the huge performance difference between TAC-KBP2015 and Hansel (discussed in Section \ref{sec:results_gap}).

\subsection{Experiment Details}
\label{sec:experiments}
\textbf{DE, TyDE and CA models} are implemented with Tensorflow \cite{abadi2016tensorflow}. 
The DE, TyDE and CA encoders all use 12 transformer encoder layers, initialized with Chinese BERT-base parameters.
The number of parameters for DE, TyDE and CA are roughly 204M, 210M, 102M.

The models are trained on a single NVIDIA V100 GPU.
We use Adam optimizer \cite{kingma2015adam} with linear weight decay and use 10\% steps for a linear warmup schedule. 
All general models are trained for 100K steps.
Training of DE and TyDE model takes approximately 30 hours.
Training CA on Wikipedia takes 16 hours, and finetuning CA on TAC-KBP2015 takes 4 hours.

We fix sequence length to be 128 tokens for both mention and entity encoder for DE and TyDE, and 256 tokens for CA. 
The batch size is 64 for DE and TyDE, and 32 for CA.
We search learning rate among [1e-5, 2e-5, 1e-4] for DE and TyDE. Following \citet{el100}, we fix 1e-5 as the learning rate for CA. 
We search learning rate among [1e-6, 5e-6] for CA-tuned. 
We search mention and entity embedding dimension $d$ within [128, 256] for DE and TyDE.
We use accuracy in validation set to make hyper-parameter choices.
Best-performing hyper-parameters are: embedding dimension $d$ is 256, learning rate is 2e-5 for DE and TyDE, and 5e-6 for CA-tuned.

\textbf{MGENRE.} 
For mGENRE's performance on \DS, We use the mGENRE model in the publicly available GENRE repository\footnote{\url{https://github.com/facebookresearch/GENRE}}. 
We do not perform any fine-tuning to its parameters. Since mGENRE uses both Wikipedia and Wikidata dumps from 2019-10-01, and Hansel-ZS include entities from Wikidata 2021-03-15, we extend mGENRE's catalog of entity names with all languages for every entity in $E_{new}$ for \DS-ZS evaluation.

\begin{table}[t]
\centering
\begin{tabular}{cc r}
\toprule
  & \bf Metric & \bf Value   \\
\midrule
\citet{tsai-roth-2016-cross}      & R@1  & 85.1      \\
\citet{sil2018neural}             & R@1  & 85.9      \\
\citet{upadhyay-etal-2018-joint}  & R@1  & 86.0      \\
\citet{zhou2019towards}           & R@1  & 85.9      \\
\citet{mgenre}                    & R@1  & \textbf{88.4}  \\
\midrule
\textbf{DE}       & R@1 & 75.2  \\  
\textbf{TyDE}     & R@1 & 76.2   \\  
\textbf{CA}       & R@1 & 81.7   \\  
\textbf{CA-tuned}       & R@1 & \underline{88.1}   \\

\midrule
\emph{AT-base} & R@1   & 73.1  \\
\emph{AT-base} & R@10  & 89.1  \\
\emph{AT-base} & R@100 & 89.4  \\
\emph{AT-ext} & R@1   & 75.3   \\
\emph{AT-ext} & R@10  & 91.1   \\
\emph{AT-ext} & R@100 & 91.5   \\
\bottomrule
\end{tabular}
\caption{Results on TAC-KBP2015 Chinese EL task. Our monolingual CA-tuned is on a par with the multi-lingual SOTA. We also report recall of our base and extended alias tables.
\label{tab:results_kbp}}
\end{table}

\subsection{Evaluation on TAC-KBP2015}
\label{sec:eval_tackbp2015}

To compare our models with prior work, we benchmark on the established TAC-KBP2015 Chinese EL task.
Note that TAC-KBP2015 was originally designed for cross-lingual EL, but still suitable as a monolingual benchmark.
Following \citet{mgenre}, we do not consider the annotated NIL entities in the dataset. We use full Chinese Wikipedia ($E_{known}$ and $E_{new}$) as our target KB\footnote{We use a Freebase API to resolve predictions to a Freebase MID, to be consistent with the dataset. When our system cannot resolve the link, it counts as a prediction error.}.
The evaluation metric is Recall@K, where R@1 is equivalent to accuracy.

Following \citet{mgenre}, we use the TAC-KBP2015 train set to extend \emph{AT-base}, denoted as \emph{AT-ext}. 
Models are trained with $E_{known}$ examples only, as described in Section \ref{sec:dataset_train}, where only \emph{AT-base} was used for generating negatives.
We further fine-tune CA on TAC-KBP2015's training set for 1 epoch, using \emph{AT-ext} to generate negatives.
The finetuned model is denoted as \emph{CA-tuned}.

We evaluate DE, TyDE and CA models, based on \emph{AT-ext}'s top-10 candidates.
\autoref{tab:results_kbp} shows evaluation results. Despite using a monolingual approach, 
our CA-tuned is on a par with the state-of-the-art model using multilingual data for training.
In particular, CA-tuned outperforms all previous cross-lingual models \cite{sil2018neural, upadhyay-etal-2018-joint,zhou2019towards}.

\textbf{Inference strategy of TyDE.} We experiment with TyDE's different typing score combinations for inference in
\autoref{tab:results_tyde}.
Combining only fine typing score, i.e. $P(e|m)sim(m,e)s_{f}(m,e)$, performs the best among different settings.
We will adopt this setting for TyDE's inference on Hansel.
TyDE improves over a standard DE with minimal added complexity.
In addition, we find that combining coarse typing score leads to performance degradation.
A possible reason is that five coarse types are not enough for entity disambiguation,
and extending to 10,000 TopSnaks encourages the learned mention and entity embeddings
to capture diverse attributes in fine types.

\begin{table}[t]
\centering
\begin{tabular}{l r}
\toprule
\textbf{Strategy}  &  \textbf{R@1}   \\
\midrule
DE                     & 75.2   \\
TyDE (sim only)        & 75.9 \\
TyDE (sim+coarse)      & 74.9 \\
TyDE (sim+fine)        & \textbf{76.2} \\
TyDE (sim+coarse+fine) & 75.1 \\
\bottomrule
\end{tabular}
\caption{Evaluations of \TYDE \space inference strategy on TAC-KBP2015. We compare multiplying similarity with coarse, fine or both typing scores.
\label{tab:results_tyde}}
\end{table}

\textbf{Error Analysis.} 
Among all R@1 errors of CA-tuned,
in 211 (20\%) cases, the gold entities do not have a Chinese Wikipedia page and do not exist in either $E_{known}$ or $E_{new}$, so our model misses these examples, 
whereas cross-lingual and multilingual models \cite{upadhyay-etal-2018-joint,mgenre} are inherently better at such examples.
545 (53\%) errors do not have the mention-entity pair in alias table's top-10 candidates, suggesting major headroom of overcoming the restriction of alias tables.
In 256 (25\%) cases, the model does not choose the correct candidate.
In 20 (2\%) cases, the freebase MIDs are not resolved to Wikidata.

\subsection{Evaluation on Hansel}
\label{sec:eval_hansel}

We set up baselines on Hansel-FS and Hansel-ZS with our models. When evaluating on \DS, we do not use dataset-specific tuning. We use \emph{AT-base} as the alias table and evaluate DE and CA based on \emph{AT-base}’s top-10 candidates.
\autoref{tab:results_hansel} shows the results on Hansel.

\begin{table*}[t]
\centering
\small
\begin{tabular}{r ccccc cccc cc} \toprule

  & \multicolumn{9}{c}{\bf In-KB}
  & \multicolumn{2}{c}{\bf With-NIL} \\
  \cmidrule(lr){2-10}  \cmidrule(lr){11-12}
  
  & \multicolumn{3}{c}{\bf AT} 
  & \multicolumn{1}{c}{\bf TyDE} 
  & \multicolumn{1}{c}{\bf CA} 
  & \multicolumn{1}{c}{\bf GEN.}
  & \multicolumn{1}{c}{\bf +margin}
  & \multicolumn{1}{c}{\bf +cand}
  & \multicolumn{1}{c}{\bf +both}
  & \multicolumn{1}{c}{\bf AT} 
  & \multicolumn{1}{c}{\bf CA+TyDE} 
  \\ 
\cmidrule(lr){2-4}  \cmidrule(lr){5-5}  \cmidrule(lr){6-6}  \cmidrule(lr){7-10}
\cmidrule(lr){11-11}
\cmidrule(lr){12-12}

\textbf{Metric} & R@1 & R@10 & R@100 & R@1  & R@1& R@1& R@1& R@1& R@1 & R@1 & R@1 \\ 
\midrule
Hansel-FS     & 0.0 & 61.1 & 63.0 & 11.7  & \textbf{46.2} & 36.6 & 35.2 & 35.2 & 35.6 & 0.0 & 44.1 \\  
Hansel-ZS     & 70.6 & 78.5 & 78.8 & 71.6  & \textbf{76.6} & 67.9\mbox{*} & 66.8\mbox{*} & 68.4\mbox{*} & 68.4\mbox{*} & 63.0 & 70.7 \\  
\bottomrule
\end{tabular}
\caption{\small 
Evaluation of our baselines, mGENRE (denoted as GEN.) and mGENRE's variants (+margin, +cand, +both) on the \DS \space dataset. 
Both datasets are challenging for the state-of-the-art MEL model, while our CA model generalizes better to few-shot and zero-shot settings. mGENRE numbers on Hansel-ZS\mbox{*}: does not follow zero-shot training constraints, but still lower than CA results.
\label{tab:results_hansel}}
\end{table*}

\textbf{Comparison with mGENRE.} We evaluate the state-of-the-art mGENRE for comparison.
From \autoref{tab:results_hansel}, the base version of mGENRE outperforms its variants with candidates and marginalization.
This may be due to the low recall of \emph{AT-base} on Hansel-FS, while the base version can recover some alias table misses.
On Hansel-FS, our CA model outperforms mGENRE by 9.6 points.
On Hansel-ZS, although mGENRE was trained on a Wikidata dump that overlaps with $E_{new}$, partially violating the zero-shot constraint, the best variant of mGENRE still under-performs CA (-8.7).

In short, CA currently achieves the best result for both zero-shot (76.6\%) and few-shot (46.2\%) slices, outperforming mGENRE by a large margin on both scenarios.
This suggests that CA is less prone to popularity bias and generalizes better to tail and emerging entities.
Large room of improvement remains on both datasets.

\textbf{Error analysis.} Among all R@1 errors of CA on Hansel-FS,
75\% errors do not have the mention-entity pair in alias table's top-10 candidates. 
For other errors, we sample 40 of them and find 40\% errors are confusion with attributes such as location and time
(e.g., "Shizhong District" in different cities and "Summer Olympics" in different years).
In 43\% cases, CA confuses general entities with specific instances
(e.g., predicts "2020 Russian constitutional referendum" while the ground truth entity is "constitutional amendment").

\textbf{With-NIL evaluation.} 
NIL entities are given a coarse type during annotation.
A NIL entity is correctly linked only if both conditions are met: 
(1) the model can predict the mention corresponds to a NIL instead of a known QID, i.e. no prior information is given whether the entity is in KB or not;
(2) the coarse type is classified correctly.
In the CA+TyDE baseline, 
we use CA to rank \emph{AT-base}'s top-10 candidates and use TyDE's coarse classification head to compute NIL type. 
A NIL output is predicted if there is no candidate with CA's output score above a threshold of 0.1. 
We combine CA's NIL judgement with TyDE's coarse typing result, 
and report the results in \autoref{tab:results_hansel} as the baseline.

\begin{table*}[t]
\centering
\begin{tabular}{r rrr rrr rrr r r} \toprule
  \multicolumn{1}{c}{\multirow{2}{*}{\textbf{Dataset}}}
  & \multicolumn{3}{c}{\bf \#Mentions} 
  & \multicolumn{3}{c}{\bf \#Distinct Mentions} 
  & \multicolumn{3}{c}{\bf \#Documents} 
  & \multicolumn{1}{c}{\multirow{2}{*}{\textbf{\#Entities}}}
  & \multicolumn{1}{c}{\multirow{2}{*}{\textbf{Domains}}}
  \\ 
\cmidrule(lr){2-4}  \cmidrule(lr){5-7}  \cmidrule(lr){8-10}  
 & In-KB & NIL & Total & In-KB & NIL & Total & In-KB & NIL & Total \\ 
\midrule
TAC-KBP2015 \cite{ji2015overview} & 8,666 & 2,400 & 11,066 & 1,246 & 1,627 & 2,869 & 166 & 146 & 166 & 840 & News, Forum \\
TAC-KBP2016 \cite{ji2016overview} & 7,115 & 1,730 & 8,845 & 1,185 & 1,080 & 2,221 & 166 & 167 & 167 & 742 & News, Forum \\
TAC-KBP2017 \cite{ji2017overview} & 7,673 & 2,573 & 10,246 & 1,218 & 1,297 & 2,421 & 167 & 167 & 167 & 796 & News, Forum \\
CLEEK \cite{zeng2020cleek} & 2,609 & 177 & 2,786 & 1,435 & 135 & 1,569 & 100 & 55 & 100 & 1,191 & News \\
\midrule
TAC-KBP2015 FS Subset & 2,072 & 316 & 2,388 & 417 & 140 & 555 & 155 & 90 & 161 & 298 & News, Forum \\
TAC-KBP2016 FS Subset & 2,255 & 581 & 2,836 & 475 & 241 & 679 & 166 & 130 & 167 & 354 & News, Forum \\
TAC-KBP2017 FS Subset & 2,583 & 1,300 & 3,883 & 486 & 464 & 877 & 163 & 159 & 167 & 350 & News, Forum \\
CLEEK FS Subset & 685 & 47 & 732 & 421 & 36 & 456 & 94 & 24 & 95 & 377 & News \\
\midrule
Hansel-FS (ours) & 3,404 & 1,856 & 5,260 &  2,654 & 1,606 & 4,097 &  3,389 & 1,850 & 5,234  & 2,720 & News, Social Media \\
Hansel-ZS (ours) & 4,208 & 507   & 4,715 &  3,981 & 468 & 4,222 &  4,200 & 507  &  4,704  & 4,046 & \makecell[r]{News, Social Media, \\ E-books, etc.} \\
\bottomrule
\end{tabular}
\caption{
\small \textbf{Comparision of existing Chinese EL datasets and the \DS \space dataset.}
We break down the number of mentions, distinct mentions and documents by whether the label is a NIL entity or inside Wikidata (In-KB). We also provide statistics of existing datasets' few-shot (FS) subsets.
\label{tab:discuss_datasets}}
\end{table*}

\section{Discussion}

In this section, with regard to each contribution listed in Section~\ref{sec:intro}, we discuss about 
the necessity of Hansel (Section \ref{sec:compare_discuss}), 
exporting annotation methods to other languages (Section \ref{sec:extend_to_new_language}), 
and the performance difference on TAC-KBP2015 and Hansel (Section \ref{sec:results_gap}).

\subsection{Comparision of Hansel and Existing Chinese EL Datasets}
\label{sec:compare_discuss}

The only 2 series of Chinese EL datasets that link to Wikidata are TAC-KBP series \cite{ji2015overview, ji2016overview, ji2017overview} and CLEEK \cite{zeng2020cleek}. \autoref{tab:discuss_datasets} summarizes the datasets' statistics and domains. Hansel sets itself apart by filling the vacancy of non-English few-shot and zero-shot datasets. 

To obtain a few-shot slice, it is intuitive to subsample existing datasets, 
i.e. removing correct AT@1 examples as the human annotation stage does. 
Although subsampling is feasible, 
the major disadvantage still exists, i.e. the lack of mention and entity diversity.
As \autoref{tab:discuss_datasets} shows, the few-shot subsets of TAC-KBP and CLEEK lack diversity due to their intrinsic features. Take TAC-KBP2017 for example, its subset has 3,883 mentions, covering only 877 different surface forms, 167 documents and 350 entities, suggesting lots of lexical repetitions across examples. In contrast, Hansel-FS has 5,260 (1.4x) mentions, covering 4,097 (5x) different surface forms, 5,234 (30x) documents and 2,720 (8x) entities. 
The diversity of Hansel-FS is rooted from our collection method, as we sample mentions from a large set of documents and avoid repetitive mentions and entities, making the dataset challenging and syntactically diverse.

For Hansel-ZS, we use the emerging entities in temporally evolving Wikidata to collect data. We apply this zero-shot setting due to its practical use. Since EL is often used in knowledge base construction and population \cite{shen2014entity, hoffart2014discovering}, this setting simulates how to link mentions to emerging entities with 2018's training data.

The TAC-KBP datasets are available for a price, but Hansel is open-source for the convenience of future research.

In conclusion, Hansel provides a comprehensive and open-source EL benchmark and cannot be substituted by simply subsampling.

\subsection{Extending to Other Languages}
\label{sec:extend_to_new_language}

To port our annotation method to a new language, 
one may re-use our Wikidata entity splits, i.e. $E_{new}$ and $E_{known}$.
Then, one may obtain an alias table by parsing language-specific Wikipedia.
With a large text corpus for the language, one may adopt our matching-based process in Section \ref{sec:dataset_fewshot} 
for a few-shot EL dataset. 
For new entities (also applicable for few-shot entities if no large corpus is available), 
one may refer to the searching-based method in Section \ref{sec:dataset_zeroshot}.
The annotators need to have expertise in the target language.

\subsection{Performance Difference}
\label{sec:results_gap}

\textbf{Difference between TAC-KBP2015 and Hansel.} 
Compared with \textbf{overall results} on TAC-KBP2015 (see \autoref{tab:results_kbp}), results on Hansel decrease across all models (see \autoref{tab:results_hansel}), suggesting the challenge of few-shot and zero-shot setting. 
For \textbf{CA and mGENRE comparison}, on TAC-KBP2015, CA has a slightly lower performance (88.1\%) than mGENRE (88.4\%), i.e. mGENRE gets 26 more correct examples than CA. 
However, the results of CA is higher than mGENRE by a large margin on both slices of Hansel.
In conclusion, CA does significantly better on tail and new entities than mGENRE, while keeping a strong performance on general entities. 
The big performance variance between mGENRE and CA on TAC-KBP2015 and Hansel-FS indicates the popularity bias issue raised by \citet{ling2021}. 
Hansel will facilitate future research that reduce such biases.

\textbf{Few-shot setting harder than zero-shot setting?} 
This observation is actually coherent with prior state-of-the-art systems. 
These EL systems achieve higher scores in zero-shot settings than few-shot settings on English EL datasets: \citet{wu2019scalable} achieve the best performance so far on the zero-shot ZESHEL \cite{zeshel} with an accuracy of 63.0\%,
but in the few-shot setting, the accuracy is only 49.0\% on the "Tail" slice of AmbER-H for fact checking \cite{ling2021}.

The reason behind this lies in the intrinsic difficulty of few-shot EL datasets. 
For each mention in Hansel-FS, 
although the ground-truth entity has appeared a few times in the training data, 
it is not the most popular entity that share the same surface form. 
Previous work sometimes refers to this few-shot setting as "overshadowed" \cite{provatorova-2021-robustness}, "tail" \cite{ling2021} or "hard" \cite{tsai-roth-2016-cross}. 
For example, the correct entity for \emph{“Michael Jordan”} in \emph{“Michael Jordan published a new paper in machine learning”} is “Michael Jordan (scientist)”, but as reported in \cite{provatorova-2021-robustness}, state-of-the-art entity linking systems GENRE \cite{genre}, REL \cite{van2020rel} and WAT \cite{piccinno2014wat} all link to the most common entity “Michael Jordan   (basketball player)”. On the other hand, for our zero-shot slice, although ground-truth entities are unseen during training, some may still be the most popular by their names, thus easier to resolve.

\section{Conclusion}
To address the popularity and language bias with entity linking (EL) datasets,
we present Hansel, a new Chinese EL benchmark consisting of two slices: 
Hansel-FS where the correct entities are not the most popular, and
Hansel-ZS where the entities are not observed in training.
We establish strong baselines on Hansel
and make the dataset and baseline models publicly available.
Along with the dataset, we propose a method to collect human-calibrated few-shot and zero-shot EL datasets, applicable for any language.
Future work on Chinese or multilingual EL may use our benchmark to test generalization over tail and emerging entities.

\bigskip
\noindent \textbf{Acknowledgments.}
We thank the valuable feedback of Yuxiang Wu, Xin Su, Yi Luan, and Yulin Chen, and the insightful suggestions of the anonymous reviewers.
This work is jointly supported by grants: Natural Science Foundation of China (No.62006061), Stable Support Program for Higher Education Institutions of Shenzhen (No.GXWD20201230155427003-20200824155011001) and Strategic Emerging Industry Development Special Funds of Shenzhen (No. JCYJ20200109113441941).

\bibliographystyle{ACM-Reference-Format}
\balance
\bibliography{custom}

\end{document}